\documentclass[runningheads]{llncs}
\usepackage[T1]{fontenc}
\usepackage{graphicx}
\usepackage{booktabs}
\usepackage[misc]{ifsym}

\usepackage{times}
\usepackage{soul}
\usepackage{url}
\usepackage[hidelinks]{hyperref}
\usepackage[utf8]{inputenc}
\usepackage[small]{caption}
\usepackage{amsmath}
\usepackage{amssymb}
\usepackage{multirow}
\usepackage{subfigure}
\usepackage{algorithm}
\usepackage{algorithmic}
\usepackage[switch]{lineno}

\newcommand{\corr}{(\Letter)}
\usepackage{mwe}
\begin{document}

\title{Bootstrap Latents of Nodes and Neighbors for \\Graph Self-Supervised Learning}

\titlerunning{Bootstrap Latents of Nodes and Neighbors}
% If the full title of your paper is short enough to also fit in the running head, you can omit the abbreviated paper title here. You can check as follows: if you comment out the \titlerunning line, something will appear in the header of all odd-numbered pages of your PDF from page 3 onward. This something is either the full title (in which case all is well), or the error message "Title Suppressed Due to Excessive Length". If this error message appears, you're going to want to provide an abbreviated title within the \titlerunning command, because if you won't do it, Springer will do it for you.

%N.B.: Author information (both in the \author{} and \authorrunning{} command) should only be present in the Camera-Ready Version of your paper. The version that you initially submit for review, ought to be double-blind. So, when initially submitting your paper, use:
% \author{Author information scrubbed for double-blind reviewing}

\author{Yunhui~Liu\inst{1,2} \and
Huaisong~Zhang\inst{3} \and
Tieke~He\inst{1,2} \corr \and
Tao~Zheng\inst{1,2} \and
Jianhua~Zhao\inst{1,2}}

% You may leave out the orcidID information, if you want to.
% Use \corr to indicate the corresponding author. Note the spacing around the \corr command. Only one author can be the corresponding author.

%N.B.: comment out the \authorrunning{} command for the double-blind version of your paper submitted for review. Later, if your paper is accepted, use the command for the Camera-Ready Version.
\authorrunning{Y. Liu et al.}
% First names are abbreviated in the running head.
% If there is one author, write 'A.L. Benjamin'.
% If there are two authors, write 'A.L. Benjamin and C.C. Broadus Jr.'
% If there are more than two authors, '[...] et al.' is used.

\institute{State Key Laboratory for Novel Software Technology, Nanjing University, Nanjing, China \email{\{lyhcloudy1225,hetieke\}@gmail.com}
\and
Software Institute, Nanjing University, Nanjing, China
\and
Tsinghua Shenzhen International Graduate School, Tsinghua University, Shenzhen, China}

\toctitle{Bootstrap Latents of Nodes and Neighbors for Graph Self-Supervised Learning}
\tocauthor{Yunhui~Liu,Huaisong~Zhang,Tieke~He,Tao~Zheng,Jianhua~Zhao}

\maketitle              % typeset the header of the contribution

\begin{abstract}
Contrastive learning is a significant paradigm in graph self-supervised learning. However, it requires negative samples to prevent model collapse and learn discriminative representations. These negative samples inevitably lead to heavy computation, memory overhead and class collision, compromising the representation learning. Recent studies present that methods obviating negative samples can attain competitive performance and scalability enhancements, exemplified by bootstrapped graph latents (BGRL). However, BGRL neglects the inherent graph homophily, which provides valuable insights into underlying positive pairs. Our motivation arises from the observation that subtly introducing a few ground-truth positive pairs significantly improves BGRL. Although we can't obtain ground-truth positive pairs without labels under the self-supervised setting, edges in the graph can reflect noisy positive pairs, i.e., neighboring nodes often share the same label. Therefore, we propose to expand the positive pair set with node-neighbor pairs. Subsequently, we introduce a cross-attention module to predict the supportiveness score of a neighbor with respect to the anchor node. This score quantifies the positive support from each neighboring node, and is encoded into the training objective. Consequently, our method mitigates class collision from negative and noisy positive samples, concurrently enhancing intra-class compactness. Extensive experiments are conducted on five benchmark datasets and three downstream task node classification, node clustering, and node similarity search. The results demonstrate that our method generates node representations with enhanced intra-class compactness and achieves state-of-the-art performance. Our implementation code is available at \url{https://github.com/Cloudy1225/BLNN}.

\end{abstract}

\section{Introduction}
Graph self-supervised learning (GSSL) is a promising paradigm for learning more informative representations without human annotations. Typically, GSSL models are pre-trained using well-designed pretext objectives, which serve as effective initializations for diverse downstream tasks \cite{GSSLSurvey}. Consequently, GSSL has made substantial advancements in graph representation learning. It offers performance, generalizability, and robustness metrics comparable to or even surpassing those of supervised methods \cite{DGI,BGRL,CSGCL}.

A major branch of GSSL is graph contrastive learning (GCL) methods \cite{GRACE,GCA}, which aim to learn representations by maximizing the agreement between two augmented samples (positive pair) while minimizing the similarities with other samples (negative pairs). The constructed negative pairs is crucial for preventing model collapse and generating discriminative representations \cite{UCL}. Consequently, current GCL methods inherently rely on increasing the quantity and quality of negative samples. This reliance not only introduces additional computational and memory costs but also leads to the class collision issue, where different samples from the same class are erroneously considered negative pairs, thereby impeding representation learning for classification \cite{TACL}. To address these issues, recent non-contrastive methods have explored the prospect of learning without negative samples \cite{CCASSG,BGRL,G-BT,AFGRL,iGCL,SGCL}. Among these methods, Bootstrapped Graph Latents (BGRL) \cite{BGRL}, derived from BYOL \cite{BYOL}, has achieved competitive performance and heightened scalability. BGRL learns node representations by using representations of one augmented view to predict another view, i.e., maximizing the similarity between the prediction and its paired target. Simultaneously, BGRL strategically leverages the asymmetry between the online branch (with gradient) and the target branch (without gradient) to alleviate model collapse. 

However, BGRL fails to account for inherent graph homophily, which indicates the phenomenon that neighboring nodes tend to share the same semantic label and thus offers valuable insights into underlying positive pairs. \emph{Why does exploiting the homophily pattern make sense?} In practice, some supervised metric learning methods \cite{SupCon,CTRR,SimPLE}, which employ architectures and objectives akin to self-supervised learning, have illustrated that introducing more ground-truth positive pairs (i.e., samples with the same label) significantly enhances representation learning for classification. Such success inspires us that mining potential positive pairs could empower the model to learn highly intra-class-compacted representations, which are more conducive to classification. Our hypothesis is validated through empirical studies in Section \ref{Sec: Motivation}. Unfortunately, unlike the supervised setting, obtaining ground-truth positive pairs is unfeasible due to the absence of labels under the self-supervised setting. But fortunately, the homophily pattern is evident in various real-world graphs \cite{Homophily}, where neighboring nodes can be seen as noisy positive pairs. Consequently, exploiting such neighbor information holds promise for graph self-supervised learning.

Based on the above analysis, we propose Bootstrap Latents of Nodes and Neighbors (BLNN) to enhance Bootstrapped Graph Latents by incorporating neighbor information. Specifically, we first expand the positive pair set with node-neighbor pairs based on the graph homophily pattern. However, although connected nodes tend to share the same label in the homophily scenario, there also exist inter-class edges, especially near the decision boundary between two classes. Treating these inter-class connected nodes as positive (i.e., false positive) pairs would inevitably compromise overall performance. To alleviate this class collision caused by false positive pairs, we further introduce an attention module to compute a supportiveness score of each neighbor representation with respect to the current view anchor node. This score serves as a soft measure of the supportiveness associated with each neighbor contributing to the current anchor node during loss computations. Basically, a higher supportiveness often stands for a higher weight to intra-class node-neighbor pairs. To this end, our BLNN incorporates soft positive node-neighbor pairs to support the anchor node for loss computations, resulting in more intra-class-compacted and discriminative node representations. The contributions of our work can be summarized as follows:

\begin{itemize}
    \item We empirically demonstrate the efficacy of introducing more ground-truth positive pairs in boosting the negative-sample-free method BGRL. And we propose exploiting the graph homophily to mining positive pairs in the absence of labels. 

    \item We expand the positive pair set with node-neighbor pairs and propose a cross-attention module to weight the contribution of each neighbor to loss computations. This approach mitigates class collision resulting from false positive node-neighbor pairs.

    \item Extensive experiments are conducted on five benchmark datasets and three downstream task node classification, node clustering, and node similarity search. The results demonstrate that our method generates node representations with enhanced intra-class compactness and achieves state-of-the-art performance.
\end{itemize}

\section{Related Work}
\subsection{Graph Self-Supervised Learning}
Recently, numerous research efforts have been devoted to graph self-supervised learning, and a branch based on multi-view learning has garnered attention owing to its superior performance. The basic idea involves ensuring consensus among multiple views derived from the same sample under different graph transformations to optimize model parameters \cite{GSSLSurvey}. A crucial aspect of these methods is the prevention of trivial solutions, where all representations converge either to a constant point (i.e., complete collapse) or to a subspace (i.e., dimensional collapse). The existing methods can be broadly classified into two groups: contrastive and non-contrastive approaches, each delineated by its strategy for mitigating model collapse.

\textbf{Contrastive-based methods} typically follow the criterion of mutual information maximization \cite{DIM}, whose objective functions involve contrasting positive pairs with negative ones. Pioneering works, such as DGI \cite{DGI} and GMI \cite{GMI}, focus on unsupervised representation learning by maximizing mutual information between node-level representations and a graph summary vector, employing the Jensen-Shannon estimator \cite{f-GAN}. MVGRL \cite{MVGRL} proposes to learn both node-level and graph-level representations by performing node diffusion and contrasting node representations to augmented graph representation. GRACE \cite{GRACE} and its variants GCA \cite{GCA}, gCooL \cite{gCooL}, CSGCL \cite{CSGCL} learn node representations by pulling together the representations of the same node in two augmented views while pushing away the representations of the other nodes in two views \cite{UCL}. Despite the success of contrastive learning on graphs, they require a large number of negative samples with carefully crafted encoders and augmentation techniques to learn discriminative representations, making them suffer seriously from heavy computation, memory overhead and class collision \cite{TACL}.

\textbf{Non-contrastive methods} discard negative samples, necessitating specialized strategies to avoid collapsed solutions. CCA-SSG \cite{CCASSG}, G-BT \cite{G-BT} and iGCL \cite{iGCL} learn augmentation invariant information while introducing feature decorrelation to capture orthogonal features and prevent dimensional collapse. BGRL \cite{BGRL}, derived from BYOL \cite{BYOL}, introduces an online network along with a target network, where the target network is updated with a moving average of the online network to avoid collapse. AFGRL \cite{AFGRL} identifies nodes as positive samples by considering both local structural information and global graph semantics, sidestepping the need for an augmented graph view and negative sampling. SGCL \cite{SGCL} uncovers the hidden factors contributing to BGRL's success and simplifies the architecture design. In this paper, we propose mining potential positive pairs from neighboring nodes to enhance BGRL.

\subsection{Generation of Positive and Negative Pairs}
There are two common approaches to generating positive and negative pairs, depending on the availability of label information. In the supervised setting, where label information is available, positive pairs consist of samples within the same class, while negative pairs comprise samples from different classes \cite{SupCon,CTRR,SimPLE}. In the self-supervised setting without label information, a typical strategy is to generate different views of the original sample via augmentation \cite{PowerOfContrast}. Here, two views of the same sample serve as positive pairs for each other, while those of different samples serve as negative pairs. However, such instance discrimination based methods inevitably a class collision issue, which means even for very similar samples, they still need to be pushed apart. 

To mitigate the class collision issue, some studies focus on mining positive pairs from nearest neighbors \cite{WCL,NNCLR,SNCLR,AFGRL} while some propose methods without negative pairs \cite{BYOL,CCASSG,BGRL,AFGRL}. In the domain of graph, AF-GCL \cite{AF-GCL} regards multi-hop neighboring nodes as potential positive pairs, utilizing well-designed similarity metrics to identify the most similar nodes as positive pairs; nevertheless, this method still necessitates a considerable number of negative pairs. AFGRL \cite{AFGRL} and HomoGCL \cite{HomoGCL} identify positive pairs by considering the local structural information and the global semantics of graphs, but they require performing time-consuming K-means clustering on the entire set of node representations to capture global semantic information. Our BLNN differs from previous work in the following three highlights: 1) BLNN, derived from BGRL \cite{BGRL}, is a non-contrastive method, eliminating the introduction of class collision arising from false negative pairs. 2) BLNN treats all one-hop node-neighbor pairs as candidate positive pairs, simplifying the selection of candidate neighbors from the K-NN search. 3) BLNN employs a cross-attention module, instead of the time-consuming K-means, to mitigate class collision caused by noisy positive node-neighbor pairs.

\section{Preliminary}
\subsection{Problem Statement}
Let $\mathcal{G} = (\mathcal{V}, \mathcal{E})$ represenst an attributed graph, where $\mathcal{V} = \{ v_1, v_2, \cdots, v_n\}$ and $\mathcal{E} \subseteq \mathcal V \times \mathcal V$ denote the node set and the edge set, respectively. The graph $\mathcal{G}$ is associated with a feature matrix $\boldsymbol{X} \in \mathbb{R}^{n \times p}$, where $\boldsymbol{x}_i \in \mathbb{R}^p$ represents the feature of $v_i$, and an adjacency matrix $\boldsymbol{A} \in \{ 0,1 \}^{n \times n}$, where $\boldsymbol{A}_{i,j} = 1$ if and only if $(v_i, v_j) \in \mathcal{E}$. During training in the self-supervised setting, no task-specific labels are provided for $\mathcal{G}$. The primary objective is to learn an embedding function $f_\theta(\boldsymbol{A}, \boldsymbol{X})$ that transforms $\boldsymbol{X}$ to $\boldsymbol{H}$, where $\boldsymbol{H} \in \mathbb{R}^{n \times d}$ and $d \ll p$. The pre-trained representations are intended to encapsulate both attribute and structure information inherent in $\mathcal{G}$ and can be easily transferable to various downstream tasks such as node classification, node clustering, and node similarity search.

\subsection{Graph Homophily}
Graph homophily suggests that neighboring nodes often belong to the same class, offering valuable prior knowledge in real-world graphs such as citation networks, co-purchase networks, or friendship networks \cite{Homophily}. A well-used metric for quantifying graph homophily is edge homophily, which is defined as the fraction of intra-class edges:
\begin{equation}
    \mathcal{H} = \frac{1}{|\mathcal{E}|} \sum_{(v_i,v_j) \in \mathcal{E}} \mathbb{I}(y_i=y_j),
\end{equation}
where $y_i$ denotes the class of $v_i$ and $\mathbb{I}$ represents the indicator function. In Table \ref{Tab: Dataset statistics}, edge homophily values for five benchmark datasets are presented. The table illustrates that the majority of edges are intra-class, indicating the potential to mine positive pairs from node-neighbor pairs.

\subsection{Bootstrapped Graph Latents}\label{Sec: BGRL}
We first introduce the pioneer work Bootstrapped Graph Latents (BGRL) \cite{BGRL}, which aims to maximize the similarity between representations of the same node generated from two different augmented graph views and employs asymmetric architectures to avoid collapsed representations. BGRL consists of three major components: 1) a random graph augmentation generator $\mathcal{T}$; 2) two asymmetric graph encoders, i.e., the online encoder $f_\theta$ and the target encoder $f_\phi$; 3) an objective function to maximize the similarity between the positive pair.

\noindent\textbf{Graph View Augmentation.}\quad
Given the adjacency matrix $\boldsymbol{A}$ and feature matrix $\boldsymbol{X}$ of a graph $\mathcal{G}$, BGRL employs feature masking and edge dropping to enhance both graph attributes and topological information (see Appendix A.3). The augmentation function $\mathcal{T}$ comprises all possible graph transformation operations, and each $t \sim \mathcal{T}$ corresponds to a specific transformation applied to graph $\mathcal{G}$.  At each training epoch, BGRL first samples two random augmentation functions $t^1 \sim \mathcal{T}$ and $t^2 \sim \mathcal{T}$, and then generates two views $\mathcal{G}^1 = (\boldsymbol{A}^1, \boldsymbol{X}^1)$ and $\mathcal{G}^2 = (\boldsymbol{A}^2, \boldsymbol{X}^2)$ based on the chosen functions.

\noindent\textbf{Node Representations Generation.}\quad
Different from the classical contrastive learning frameworks with a shared graph encoder, BGRL employs two asymmetric graph encoders to avoid representation collapse. The online encoder $f_\theta$ generates an online representations from the first augmented graph, $\boldsymbol{H}^1 = f_\theta(\boldsymbol{A}^1, \boldsymbol{X}^1)$. Similarly, the target encoder $f_\phi$ produces a target representation of the second augmented graph, $\boldsymbol{H}^2 = f_\phi(\boldsymbol{A}^2, \boldsymbol{X}^2)$. The online representation is then input into a node-level predictor, $p_\theta$ (implemented as a MLP), which produces a prediction of the target representation, $\boldsymbol{Z}^1 = p_\theta(\boldsymbol{H}^1)$.

\noindent\textbf{Positive Pair Similarity Maximization.}\quad
The learning process of BGRL centers around maximizing the cosine similarity between the predicted target representations $\boldsymbol{Z}^1$ and the true target representations $\boldsymbol{H}^2$, i.e., positive pairs. The objective function is defined as
\begin{equation}
    \mathcal{L}_{BGRL} = - \frac{1}{n} \sum_{i=1}^{n} \frac{\boldsymbol{z}^1_i \cdot \boldsymbol{h}^2_i}{\parallel\boldsymbol{z}^1_i\parallel \parallel\boldsymbol{h}^2_i\parallel},\label{Eq: BGRL Objective}
\end{equation}
where $(\cdot)$ denotes the dot production, and $\parallel\cdot\parallel$ represents the $\ell_2$ normalization. Notably, only the online encoder parameters $\theta$ are updated with respected to the gradients from the objective function while the target encoder parameters $\phi$ are updated as an exponential moving average (EMA) of $\theta$ with a decay rate $t$, i.e., $\phi = t \phi + (1-t) \theta$. Therefore, BGRL utilizes the outputs from the ensemble-optimized parameters as targets, progressively enhancing the model in a step-by-step fashion, an approach commonly known as bootstrapping.

\begin{figure*}[h]
\centerline{\includegraphics[width=1.\linewidth]{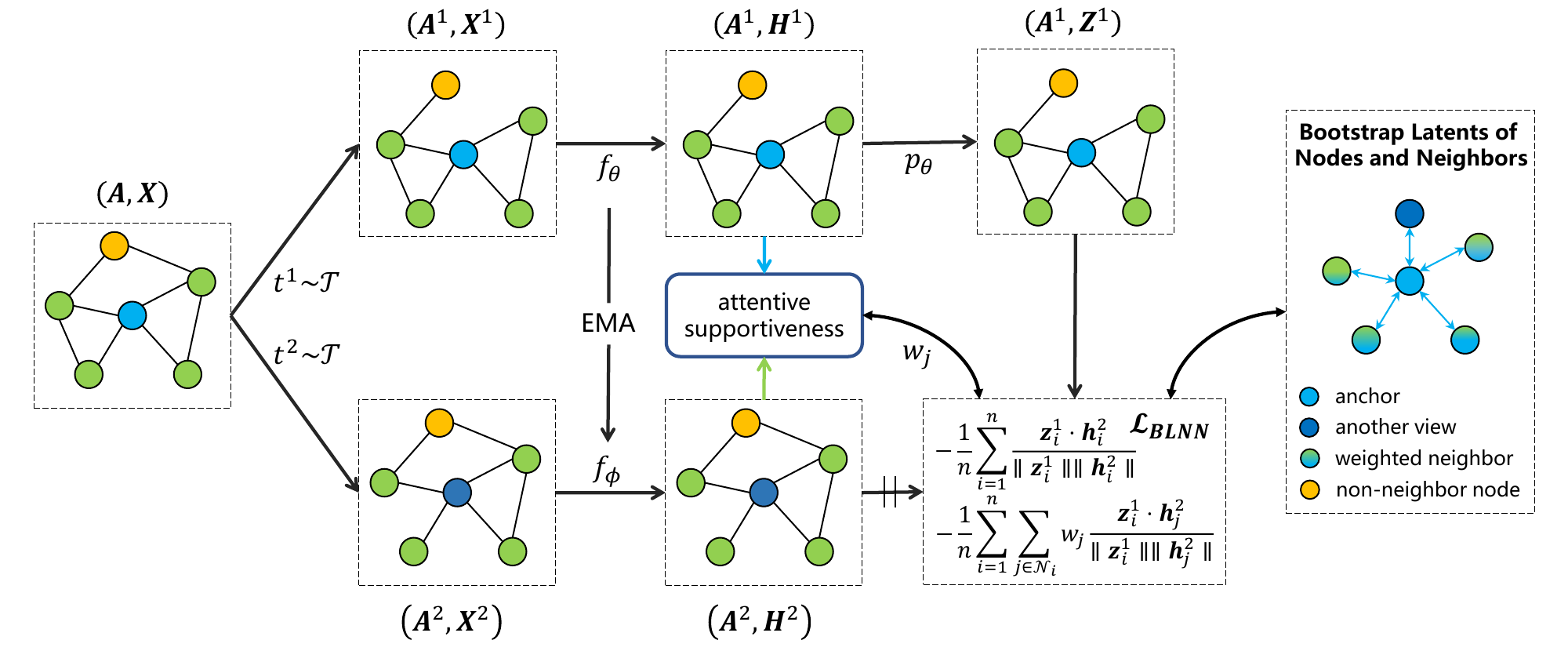}}\caption{Overview of our proposed BLNN method. Given a graph, we first generate two different views using augmentations $t^1,t^2$. From these, we use encoders $f_{\theta}, f_\phi$ to form online and target node representations $\boldsymbol{H}^1, \boldsymbol{H}^2$. They are then fed into the attention module to compute the supportiveness $w_j$ of the neighbor $v_j$ w.r.t. the anchor node $v_i$. The predictor $p_\theta$ uses $\boldsymbol{H}^1$ to form a prediction $\boldsymbol{Z}^1$ of the target $\boldsymbol{H}^2$. The final objective is computed as a combination of the alignment of node-itself pairs and the supportiveness-weighted alignment of node-neighbor pairs. Note that the alignment is achieved by maximizing the cosine similarity between corresponding rows of $\boldsymbol{Z}^1$ and $\boldsymbol{H}^2$, flowing gradients only through $\boldsymbol{Z}^1$. The target parameters $\phi$ are updated as an exponentially moving average of $\theta$.} \label{Fig: Overview}
\end{figure*}

\section{Methodology}
In this section, we present an overview of the proposed BLNN, as depicted in Figure \ref{Fig: Overview}. In Section \ref{Sec: Motivation}, we empirically analyze our motivation to introduce more ground-truth positive pairs from node-neighbor pairs for graph self-supervised learning. Then, we describe how to mine high-confidence positive information from node-neighbor pairs in Section \ref{Sec: BLNN}.

\subsection{Motivation}\label{Sec: Motivation}
As discussed in the introduction, some supervised metric learning methods \cite{SupCon,CTRR,SimPLE}, which employ architectures and objectives similar to self-supervised learning, have shown that introducing more ground-truth positive pairs significantly enhances representation learning for classification. Such success inspires us that mining potential positive pairs could empower BGRL to learn highly intra-class-compacted representations, which are more conducive to classification. 

\begin{figure}[h]
    \centering
    \subfigure[WikiCS]{\includegraphics[width=0.325\linewidth]{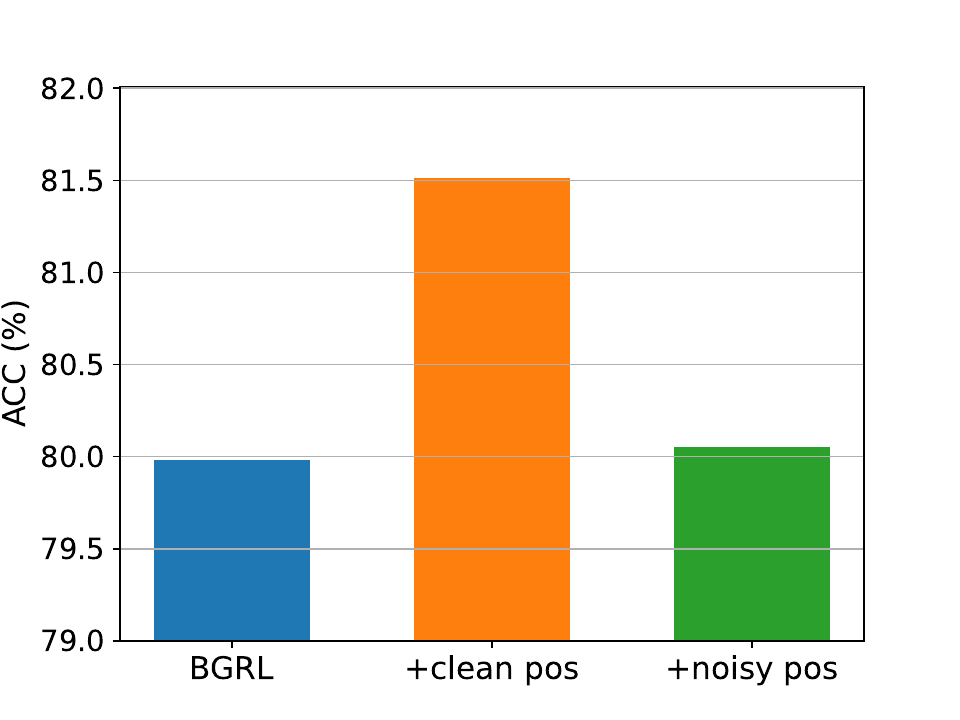}}  
    \subfigure[Computer]{\includegraphics[width=0.325\linewidth]{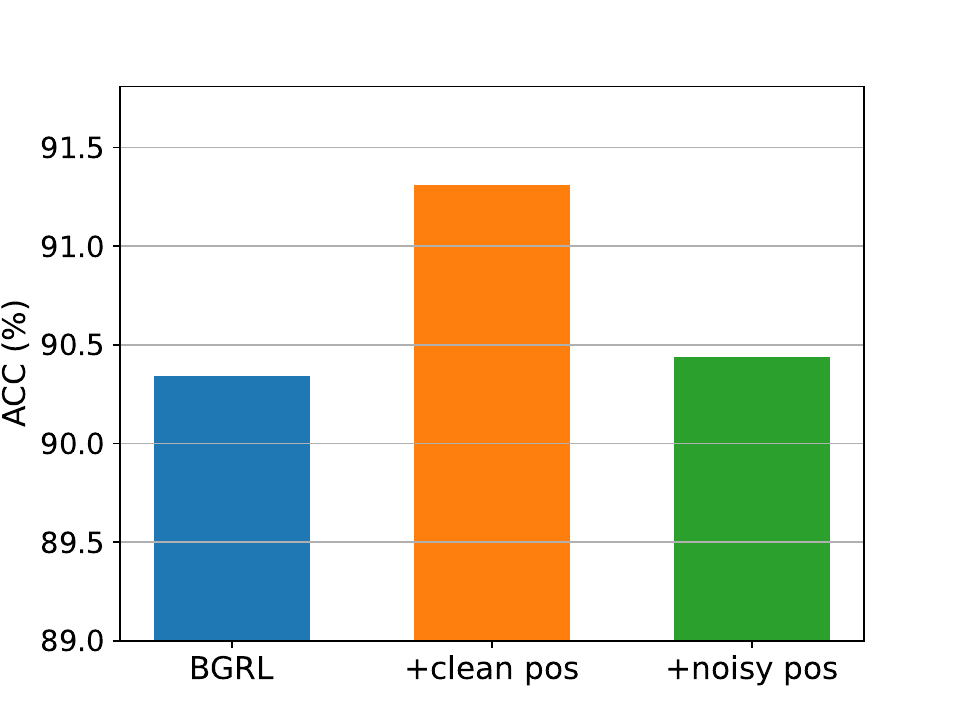}}
    \subfigure[CS]{\includegraphics[width=0.325\linewidth]{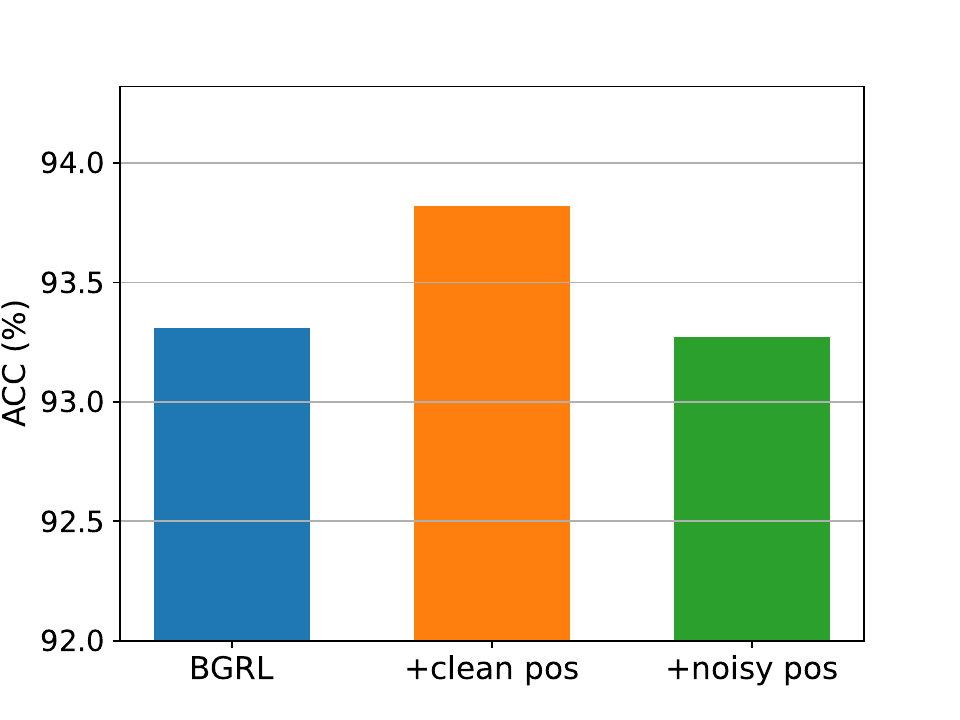}}
    \caption{Empirical studies on WikiCS, Computer and CS. ``noisy pos" indicates raw node-neighbor pairs in the input graph, while ``clean pos" indicates clean node-neighbor pairs that all are intra-class pairs.}
    \label{Fig: Empirical Studies}
\end{figure}

\textbf{Empirical Analysis.}\quad
To verify our hypothesis, we conduct experiments by incorporating a small subset of the whole ground-truth positive pair set from an oracle perspective and assessing its influence on classification. According to the graph homophily, neighboring nodes often share the same class. Therefore, we first treat all node-neighbor pairs as noisy candidate positive pairs. Subsequently, we manually filter out inter-class pairs, retaining only the intra-class pairs as the clean positive pairs. We then extend the objective function Eq.(\ref{Eq: BGRL Objective}) with an additional alignment of above intra-class node-neighbor pairs to train BGRL. Figure \ref{Fig: Empirical Studies} illustrates the results of node classification across three datasets, revealing two key observations: 1) The incorporation of clean positive node-neighbor pairs consistently and significantly improves classification performance. 2) However, simply treating raw node-neighbor pairs as ground-truth positive pairs yields only marginal improvement or even performance degradation, as raw node-neighbor pairs include inter-class pairs, which would cause class collision. 

Based on the above observations, we propose to enhance BGRL using two key strategies: 1) expanding the positive pair set with node-neighbor pairs; 2) mitigating class collision caused by false positive node-neighbor pairs via a cross-attention weighting module.

\subsection{Bootstrap Latents of Nodes and Neighbors}\label{Sec: BLNN}
Motivated by the observations presented in Section \ref{Sec: Motivation}, we introduce Bootstrap Latents of Nodes and Neighbors (BLNN) to enhance Bootstrapped Graph Latents (BGRL).  We follow the BGRL framework illustrated in Section \ref{Sec: BGRL}.

\noindent\textbf{Objective Function.}\quad
Our BLNN first treats node-neighbor pairs as candidate positive pairs, leveraging the neighbor set $\mathcal{N}_i$ to support the anchor node $v_i$. Subsequently, it introduces an adaptive measurement of supportiveness through a cross-attention module to mitigate class collision resulting from false positive node-neighbor pairs. Specifically, for each neighbor $v_j \in \mathcal{N}_i$, we input its target representation $\boldsymbol{h}^2_j$ and the anchor's online representation $\boldsymbol{h}^1_i$ into the attention module for cross-attention computations. This attention module predicts a supportiveness value $w_j$, which we use to adjust the contribution of $\boldsymbol{h}^2_j$ to the anchor's prediction $\boldsymbol{z}^1_i$ during training. The loss function of our BLNN can be written as:
\begin{equation}
\begin{aligned}
\mathcal{L}_{BLNN} = &- \underbrace{\frac{1}{n} \sum_{i=1}^{n} \frac{\boldsymbol{z}^1_i \cdot \boldsymbol{h}^2_i}{\parallel\boldsymbol{z}^1_i\parallel \parallel\boldsymbol{h}^2_i\parallel}}_{\text{Bootstrap Latents of Nodes}} \\
&- \underbrace{\frac{1}{n} \sum_{i=1}^{n} \sum_{j \in \mathcal{N}_i} w_j \frac{\boldsymbol{z}^1_i \cdot \boldsymbol{h}^2_j}{\parallel\boldsymbol{z}^1_i\parallel \parallel\boldsymbol{h}^2_j\parallel}}_{\text{Bootstrap Latents of Neighbors}}. \label{Eq: Objective Function}
\end{aligned}
\end{equation}

\noindent\textbf{Attention Weighting.}\quad
The attention module, which softly measure the positiveness of node-neighbor pairs, simply consists of a cross-attention operator, and a softmax activation. Formally, given the anchor's online representation $\boldsymbol{h}^1_i$ and its neighboring node's target representation $\boldsymbol{h}^2_j$, the supportiveness score can be computed as:
\begin{equation}
    w_j = \operatorname{softmax}_j(e_{ij})=\frac{\exp (e_{ij} / \tau)}{\sum_{k \in \mathcal{N}_i} \exp (e_{ik} / \tau)}, \label{Eq: Attention Weighting}
\end{equation}
where $e_{ij} = \boldsymbol{h}^1_i \cdot \boldsymbol{h}^2_j / \parallel\boldsymbol{h}^1_i\parallel \parallel\boldsymbol{h}^2_j\parallel$ is the cosine similarity between $\boldsymbol{h}^1_i$ and $\boldsymbol{h}^2_j$ and $\tau$ is a temperature parameter. This attention module assigns higher weights to ground-truth positive node-neighbor pairs than false positive node-neighbor pairs, thus mitigating class collision caused by aligning false node-neighbor pairs. 

\noindent\textbf{Comparison with BGRL.}\quad
Our BLNN enhances BGRL by introducing potential positive node-neighbor pairs in the absence of ground-truth labels. It inherits BGRL's advantages, such as the negative-free property, which naturally address class collision caused by false negative pairs. Different from the original BGRL framework, which aligns only augmented views with the anchor node, the cross-attention design in BLNN enriches the diversity of positive nodes to support the anchor node in a soft and adaptive manner. This design empowers us to leverage more positive pairs, enhancing intra-class compactness. Additionally, the computations for supportiveness scores and node-neighbor alignment loss exhibit a time complexity linear with the number of edges $\mathcal{O}(|\mathcal{E}|)$. Given the sparsity of real-world graphs, i.e., $\mathcal{O}(|\mathcal{E}|) << \mathcal{O}(|\mathcal{V}|^2)$, such complexity increase compared to BGRL is acceptable and our model maintains lower time complexity than contrastive learning baselines \cite{GRACE,GCA,COSTA,gCooL}.

\begin{algorithm}
    \caption{Bootstrap Latents of Nodes and Neighbors}
    \label{Alg: algorithm}
    \textbf{Input}: $\mathcal{G} = (\boldsymbol{A}, \boldsymbol{X})$\\
    \textbf{Parameter}: Temperature $\tau$, BGRL-related hyperparameters\\
    \textbf{Output}: The graph encoder $f_\theta$
    \begin{algorithmic}[1] %[1] enables line numbers
        \STATE Initialize model parameters;
        \WHILE{not converge}
        \STATE Sample two augmentation functions $t^1, t^2 \sim \mathcal{T}$;
        \STATE Generate augmented views $(\boldsymbol{A}^1, \boldsymbol{X}^1), (\boldsymbol{A}^2, \boldsymbol{X}^2)$;
        \STATE Obtain online representations $\boldsymbol{H}^1=f_\theta(\boldsymbol{A}^1, \boldsymbol{X}^1)$;
        \STATE Obtain target representations $\boldsymbol{H}^2=f_\phi(\boldsymbol{A}^2, \boldsymbol{X}^2)$;
        \STATE Compute positiveness scores of node-neighbor pairs via Eq. (\ref{Eq: Attention Weighting});
        \STATE Predict the target representations $\boldsymbol{Z}^1=p_\theta(\boldsymbol{H}^1)$;
        \STATE Calculate the objective function via Eq. (\ref{Eq: Objective Function});
        \STATE Update the parameters of $f_\theta, p_\theta$ via SGD;
        \STATE Update the parameters of $f_\phi$ via an EMA of $f_\theta$;
        \ENDWHILE
        \STATE \textbf{return} $f_\theta$.
    \end{algorithmic}
\end{algorithm}

\section{Experiments}
In this section, we design the experiments to evaluate our proposed BLNN and answer the following research questions. \textbf{RQ1}: Does BLNN outperform existing baseline methods on node classification, node clustering, and node similarity search? \textbf{RQ2}: How does each component of BLNN benefit the performance? \textbf{RQ3}: Can the supportiveness score measure the positiveness of node-neighbor pairs? \textbf{RQ4}: Is BLNN sensitive to the hyperparameter $\tau$? \textbf{RQ5}: How to intuitively understand BLNN can enhance intra-class compactness of learned representations?

\subsection{Experiment Setup}
\noindent\textbf{Datasets.}\quad
We adopt five publicly available real-world benchmark datasets, including one reference network WikiCS \cite{WikiCS}, two co-purchase networks Photo, Computer \cite{Amazon}, and two co-authorship networks CS, Physics \cite{Amazon} to conduct the experiments throughout the paper. The statistics of the datasets are provided in Table \ref{Tab: Dataset statistics}. More details can be found in Appendix A.1.

\begin{table}
    \centering
    \setlength{\tabcolsep}{4.5pt}
    \caption{Dataset statistics. $\mathcal{H}$ is the fraction of intra-class node-neighbor pairs.}\label{Tab: Dataset statistics}
    \begin{tabular}{lccccc}
        \toprule
        Dataset      & \#Nodes    & \#Edges   & \#Feats  & \#Classes   & $\mathcal{H}$ (\%)   \\
        \midrule
        WikiCS       & 11,701     & 431,726   & 300         & 10          & 65.47 \\
        Photo        & 7,650      & 238,163   & 745         & 8           & 82.72 \\
        Computer     & 13,752     & 491,722   & 767         & 10          & 77.72 \\
        CS           & 18,333     & 163,788   & 6,805       & 15          & 80.81 \\
        Physics      & 34,493     & 495,924   & 8,415       & 5           & 93.14 \\
        \bottomrule
    \end{tabular}
\end{table}

\noindent\textbf{Baselines.}\quad
We compare BLNN with a variety of baselines, including supervised methods MLP, GCN \cite{GCN}, and GAT \cite{GAT}; contrastive methods DGI \cite{DGI}, MVGRL \cite{MVGRL}, GRACE \cite{GRACE}, GCA \cite{GCA}, AF-GCL \cite{AF-GCL}, COSTA \cite{COSTA}, FastGCL \cite{FastGCL}, gCooL \cite{gCooL}, ProGCL \cite{ProGCL}, and CGKS \cite{CGKS}; non-contrastive methods CCA-SSG \cite{CCASSG}, G-BT \cite{G-BT}, AFGRL \cite{AFGRL}, GraphMAE \cite{GraphMAE}, and BGRL \cite{BGRL}. All the baseline results are taken from previously published papers. And brief introductions of the baselines can be found in Appendix A.2.

\noindent\textbf{Evaluation Protocol.}\quad
We evaluate BLNN on three tasks, i.e., node classification, node clustering and node similarity search. We first train the model in an unsupervised manner. For node classification, we use the learned representations to train and test a simple logistic regression classifier with twenty 1:1:8 train/validation/test random splits (twenty public splits for WikiCS) \cite{BGRL}. We apply K-means to the learned representations, initializing the cluster numbers with fixed values. For node similarity search, we use pairwise cosine similarity to identify nearest node neighbors \cite{AFGRL}. Evaluations are conducted at every $250$ epochs, and we report the best results \cite{BGRL,AFGRL}.

\noindent\textbf{Metrics.}\quad
Following AFGRL \cite{AFGRL}, we use accuracy for node classification, normalized mutual information (NMI) and homogeneity (Hom.) for node clustering. For node similarity search, we introduce S@$k$, which is average ratio among the $k$ nearest neighbors sharing the same label as the query node. Formulas of these metrics can be found in Appendix A.4.

\noindent\textbf{Implementation Details.}\quad
Since our BLNN is derived from BGRL, we implement BLNN based on the official code\footnote{\url{https://github.com/nerdslab/bgrl}} of BGRL. To ensure a fair comparison, all BGRL-related hyperparameters are the same as those specified in the original BGRL paper. We perform a grid-search on the introduced temperature hyperparameter $\tau$. All experiments are conducted on a 32GB V100 GPU. Our implementation code is available at \url{https://github.com/Cloudy1225/BLNN}. More details can be found in Appendix A.5.

\begin{table}[!ht]
    \centering
    \setlength{\tabcolsep}{5pt}
    \caption{Node classification results measured by accuracy along with standard deviations. The baseline results are taken from previously published papers. `-' denotes the absence of the result in the original paper. The \emph{Input} column illustrates the data used in the training stage, and $\boldsymbol{Y}$ denotes labels.}
    \label{Tab: Node Classification}
    \begin{tabular}{llccccc}
        \toprule
         Method    & Input      & WikiCS & Photo & Computer & CS & Physics   \\
        \midrule
    
         MLP       & $\boldsymbol{X}, \boldsymbol{Y}$                    & 71.98±0.00 & 78.53±0.00 & 73.81±0.00 & 90.37±0.00 & 93.58±0.00 \\
         GCN       & $\boldsymbol{A}, \boldsymbol{X}, \boldsymbol{Y}$    & 77.19±0.12 & 92.42±0.22 & 86.51±0.54 & 93.03±0.31 & 95.65±0.16 \\
         GAT       & $\boldsymbol{A}, \boldsymbol{X}, \boldsymbol{Y}$    & 77.65±0.11 & 92.56±0.35 & 86.93±0.29 & 92.31±0.24 & 95.47±0.15 \\
        \midrule
    
         DGI       & $\boldsymbol{A}, \boldsymbol{X}$                    & 78.25±0.56 & 91.69±1.07 & 87.98±0.81 & 92.15±0.63 & 94.51±0.52 \\
         MVGRL     & $\boldsymbol{A}, \boldsymbol{X}$                    & 77.57±0.46 & 92.04±0.98 & 87.39±0.92 & 92.11±0.12 & 95.33±0.03 \\
         GRACE     & $\boldsymbol{A}, \boldsymbol{X}$                    & 78.64±0.33 & 92.46±0.18 & 88.29±0.11 & 92.17±0.04 & 95.26±0.22 \\
         GCA       & $\boldsymbol{A}, \boldsymbol{X}$                    & 78.35±0.05 & 92.53±0.16 & 87.85±0.31 & 93.10±0.01 & 95.68±0.05 \\
         AF-GCL    & $\boldsymbol{A}, \boldsymbol{X}$                    & 79.01±0.51 & 92.49±0.31 & 89.68±0.19 & 91.92±0.10 & 95.12±0.15 \\
         COSTA     & $\boldsymbol{A}, \boldsymbol{X}$                    & 79.12±0.02 & 92.56±0.45 & 88.32±0.03 & 92.94±0.10 & 95.60±0.02 \\
         FastGCL   & $\boldsymbol{A}, \boldsymbol{X}$                    & 79.20±0.07 & 92.91±0.07 & 89.35±0.09 & 92.71±0.07 & 95.53±0.02 \\
         gCooL     & $\boldsymbol{A}, \boldsymbol{X}$                    & 78.74±0.04 & 93.18±0.12 & 88.85±0.14 & 93.32±0.02 & - \\
         ProGCL    & $\boldsymbol{A}, \boldsymbol{X}$                    & 78.68±0.12 & 93.30±0.09 & 89.28±0.15 & 93.51±0.06 & - \\
         CGKS      & $\boldsymbol{A}, \boldsymbol{X}$                    & 79.20±0.10 & 92.40±0.10 & 88.50±0.20 & 93.00±0.20 & - \\
        \midrule
    
         CCA-SSG   & $\boldsymbol{A}, \boldsymbol{X}$                    & 79.08±0.53 & 93.14±0.14 & 88.74±0.28 & 93.32±0.22 & 95.38±0.06 \\
         G-BT      & $\boldsymbol{A}, \boldsymbol{X}$                    & 76.83±0.73 & 92.46±0.35 & 87.93±0.36 & 92.91±0.25 & 95.25±0.13 \\
         AFGRL     & $\boldsymbol{A}, \boldsymbol{X}$                    & 77.62±0.49 & 93.22±0.28 & 89.88±0.33 & 93.27±0.17 & 95.69±0.10 \\
         GraphMAE  & $\boldsymbol{A}, \boldsymbol{X}$                    & 79.54±0.58 & 92.98±0.35 & 89.88±0.10 & 93.08±0.17 & 95.40±0.06 \\
         BGRL      & $\boldsymbol{A}, \boldsymbol{X}$                    & 79.98±0.10 & 93.17±0.30 & 90.34±0.19 & 93.31±0.13 & 95.73±0.05 \\
         BLNN      & $\boldsymbol{A}, \boldsymbol{X}$                    & \textbf{80.48±0.52} & \textbf{93.54±0.23} & \textbf{91.02±0.23} & \textbf{93.61±0.15} & \textbf{95.86±0.10} \\
        \bottomrule
    \end{tabular}
\end{table}

\subsection{Experiment Results}
\noindent\textbf{Performance Analysis (RQ1).}\quad
The experimental results of node classification are presented in Table \ref{Tab: Node Classification}, revealing that our BLNN outperforms both self-supervised and even supervised baselines. This superiority can be attributed to two primary factors: 1) The pioneering BGRL of BLNN can effectively learn discriminative node representations, achieving competitive performance. 2) BLNN introduces additional potential positive pairs, enhancing the intra-class compactness of representations learned by BGRL. Node clustering results are detailed in Table \ref{Tab: Node Clustering}, demonstrating BLNN's superior performance across four datasets, except Physics. Notably, BLNN exhibits significant improvement over BGRL, especially on WikiCS, Computer and Physics, with an increase ranging from $5\%$ to $8\%$. These enhancements underscore the effectiveness of incorporating positive node-neighbor pairs to generate more intra-class compact representations. Table \ref{Tab: Similarity Search} illustrates the node similarity search results, with BLNN demonstrating the best performance. This outcome aligns with expectations, as BLNN is designed to softly pull together representations of nodes and their neighbors, where neighboring nodes often share the same label in graphs.

\begin{table*}[!ht]
  \begin{center}
  \setlength{\tabcolsep}{5pt}
  \caption{Performance on node clustering. The baseline results are taken from the published AFGRL paper.}\label{Tab: Node Clustering}
  \begin{tabular}{c|cc|cc|cc|cc|cc}
  \bottomrule
  Dataset      & \multicolumn{2}{c|}{WikiCS}      & \multicolumn{2}{c|}{Photo}      & \multicolumn{2}{c|}{Computer}      & \multicolumn{2}{c|}{CS}         & \multicolumn{2}{c}{Physics}   \\ \hline
  Metric      & NMI           & Hom.            & NMI           & Hom.                & NMI           & Hom.              & NMI           & Hom.                 & NMI           & Hom.            \\ \hline
  GRACE       & 42.82    & 44.23      & 65.13    & 66.57         & 47.93    & 52.22       & 75.62    & 79.09         & -    &  -    \\
  GCA         & 33.73    & 35.25      & 64.43    & 65.75         & 52.78    & 58.16       & 76.20    & 79.65         & -    &  -    \\ 
  AFGRL       & 41.32    & 43.07      & 65.63    & 67.43         & 55.20    & 60.40       & 78.59    & 81.61         & \textbf{72.89}    &  \textbf{73.54}    \\ 
  BGRL        & 39.69    & 41.56      & 68.41    & 70.04         & 53.64    & 58.69       & 77.32    & 80.41         & 55.68   &  60.18  \\ 
  BLNN        & \textbf{47.17}    & \textbf{49.11}      & \textbf{71.05}    & \textbf{72.18}         & \textbf{58.79}    & \textbf{64.33}       & \textbf{78.97}    & \textbf{82.08}         & 62.41    &  67.39    \\ 
  \toprule
  \end{tabular}
  \end{center}
\end{table*}

\begin{table*}[!ht]
  \begin{center}
  \setlength{\tabcolsep}{5pt}
  \caption{Performance on node similarity search. The baseline results are taken from the published AFGRL paper.}\label{Tab: Similarity Search}
  \begin{tabular}{c|cc|cc|cc|cc|cc}
  \bottomrule
  Dataset      & \multicolumn{2}{c|}{WikiCS}      & \multicolumn{2}{c|}{Photo}      & \multicolumn{2}{c|}{Computer}      & \multicolumn{2}{c|}{CS}         & \multicolumn{2}{c}{Physics}   \\ \hline
  Metric      & S@5           & S@10            & S@5           & S@10                & S@5           & S@10              & S@5           & S@10                 & S@5           & S@10            \\ \hline
  GRACE       & 77.54    & 76.45      & 91.55    & 91.06         & 87.38    & 86.43       & 91.04    & 90.59         & -    &  -    \\
  GCA         & 77.86    & 76.73      & 91.12    & 90.52         & 88.26    & 87.42       & 91.26    & 91.00         & -    &  -    \\ 
  AFGRL       & 78.11    & 76.60      & 92.36    & 91.73         & 89.66    & 88.90       & 91.80    & 91.42         & 95.25    &  94.86    \\ 
  BGRL        & 77.39    & 76.17      & 92.45    & 91.95         & 89.47    & 88.55       & 91.12    & 90.86         & 95.04    &  94.64  \\ 
  BLNN        & \textbf{80.27}    & \textbf{79.04}      & \textbf{92.61}    & \textbf{91.96}         & \textbf{89.91}    & \textbf{89.12}       & \textbf{91.90}    & \textbf{91.59}         & \textbf{95.39}    & \textbf{95.01}    \\ 
  \toprule
  \end{tabular}
  \end{center}
\end{table*}

\noindent\textbf{Ablation Studies (RQ2).}\quad
To verify the benefit of each component of BLNN, we conduct ablation studies with different variants of BGRL: BGRL with raw nosiy node-neighbor pairs ($\text{BGRL}_{\text{noisy}}$), BGRL with clean node-neighbor pairs ($\text{BGRL}_{\text{clean}}$), and our proposed BLNN (BGRL with supportiveness-weighted node-neighbor pairs). Results are reported in Table \ref{Tab: Ablation Study}. We can find that simply treating raw node-neighbor pairs as ground-truth positive pairs results in only marginal improvement or even performance degradation, as raw node-neighbor pairs include inter-class pairs, which would cause class collision. Our supportiveness weighting strategy, implemented through an attention module, effectively mitigates this class collision, yielding superior performance. However, there is still a gap between our BLNN and the ideal solution $\text{BGRL}_{\text{clean}}$, which necessitates the availability of all labels. These results further confirm our motivation described in Section \ref{Sec: Motivation}.

\begin{table}
    \centering
    \caption{Ablation study on node classification.}
    \label{Tab: Ablation Study}
    \setlength{\tabcolsep}{5pt}
    \begin{tabular}{lccccc}
        \toprule
        Variant      & WikiCS & Photo & Computer & CS & Physics   \\
        \midrule
        BGRL         & 79.98 & 93.17 & 90.34 & 93.31 & 95.73 \\
        BLNN & 80.48 & 93.54 & 91.02 & 93.61 & 95.86 \\
        $\text{BGRL}_{\text{noisy}}$           & 80.05 & 93.33 & 90.44 & 93.27 & 95.59 \\
        $\text{BGRL}_{\text{clean}}$     & 81.51 & 93.66 & 91.31 & 93.92 & 95.98 \\
        \bottomrule
    \end{tabular}
\end{table}

\begin{figure}[h]
    \centering
    \subfigure[all node-neighbor pairs]{\includegraphics[width=0.49\linewidth]{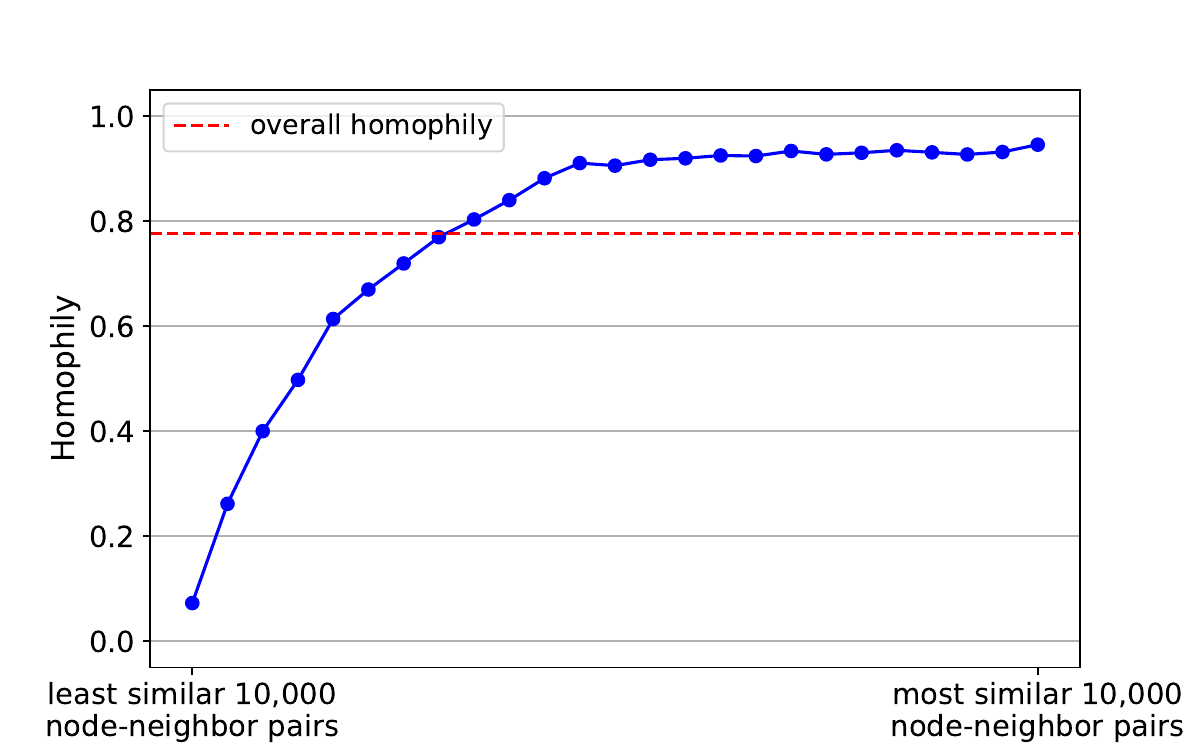}}  
    \subfigure[anchor-neighbor pairs]{\includegraphics[width=0.49\linewidth]{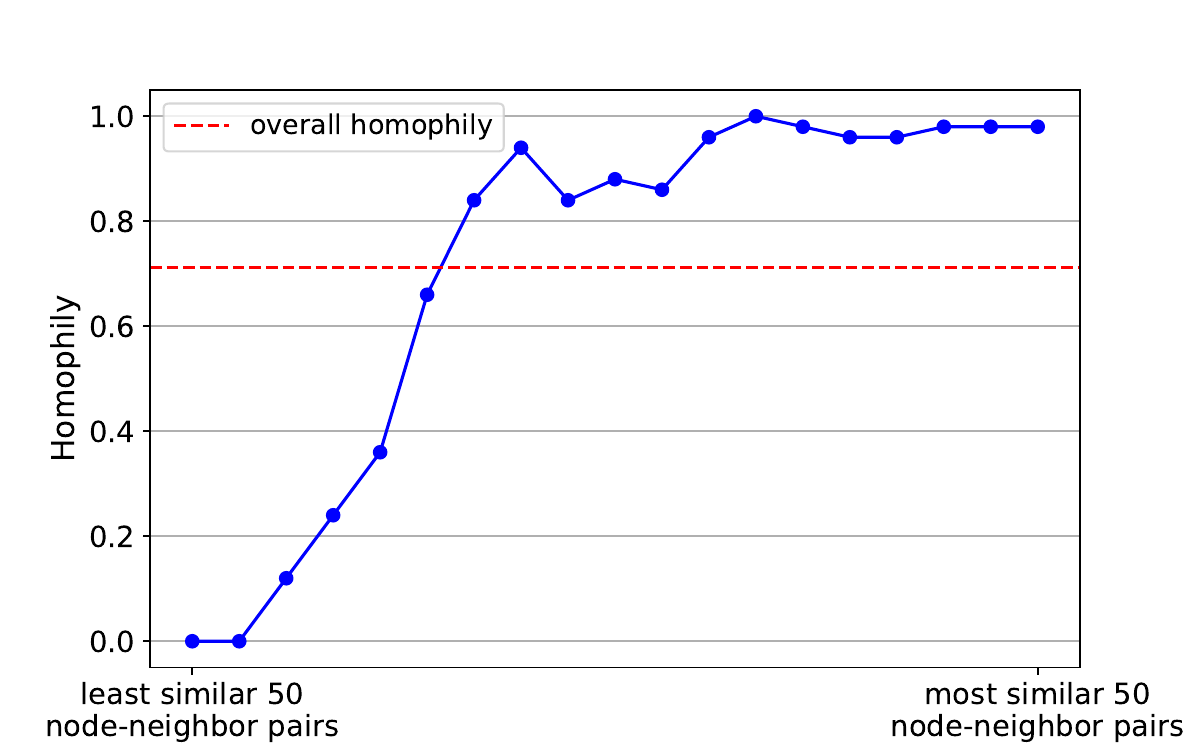}}
    \caption{Case study to verify the efficacy of our attention module.}
    \label{Fig: Case Study}
\end{figure}

\noindent\textbf{Case Study (RQ3).}\quad
Our attention module is implemented based on cosine similarities of node-neighbor pairs and is expected to assign higher weights to true positive node-neighbor pairs than false positive pairs. Here, we conduct a twofold case study on Computer to verify that: 1) node-neighbor pairs with higher cosine similarity tend to share the same label; 2) our attention module indeed assigns higher weights to true positive node-neighbor pairs. We first sort all node-neighbor pairs based on the learned cosine similarity and then divide them into intervals of size $10,000$ to compute the homophily in each interval.  As shown in Figure \ref{Fig: Case Study}(a), the cosine similarity effectively estimates the probability of neighbor nodes being positive, with more similar node-neighbor pairs exhibiting larger homophily, which validates the efficacy of leveraging cosine similarity in our attention module. Moreover, we select an anchor node with $949$ neighbors, sorting all anchor-neighbor pairs according to the supportiveness weights predicted by the attention module. We also partition them into intervals of size $50$ to calculate homophily within each interval. As shown in Figure \ref{Fig: Case Study}(b), our attention module generally assigns higher weights to true positive node-neighbor pairs compared to false positive pairs.

\begin{figure}[h]
    \centering
    \subfigure[Photo]{\includegraphics[width=0.325\linewidth]{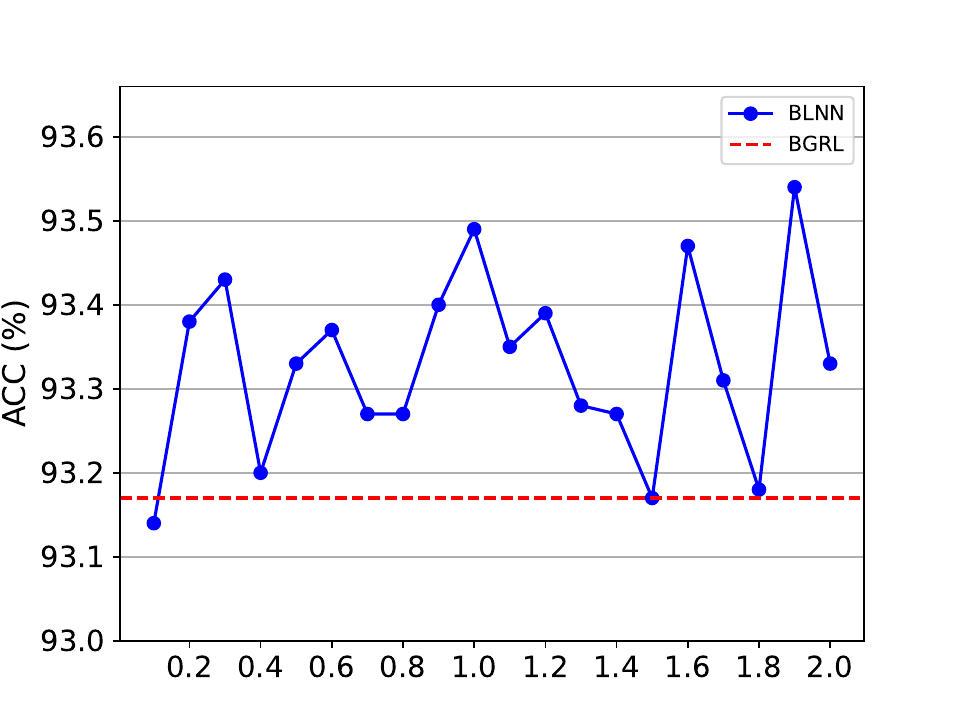}}  
    \subfigure[Computer]{\includegraphics[width=0.325\linewidth]{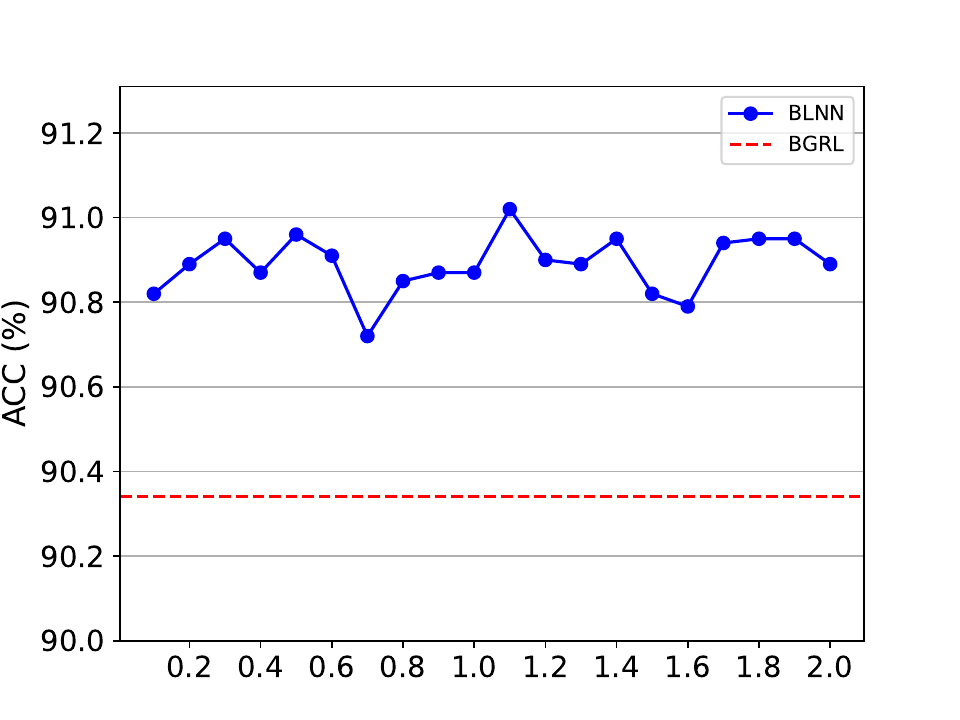}}  
    \subfigure[CS]{\includegraphics[width=0.325\linewidth]{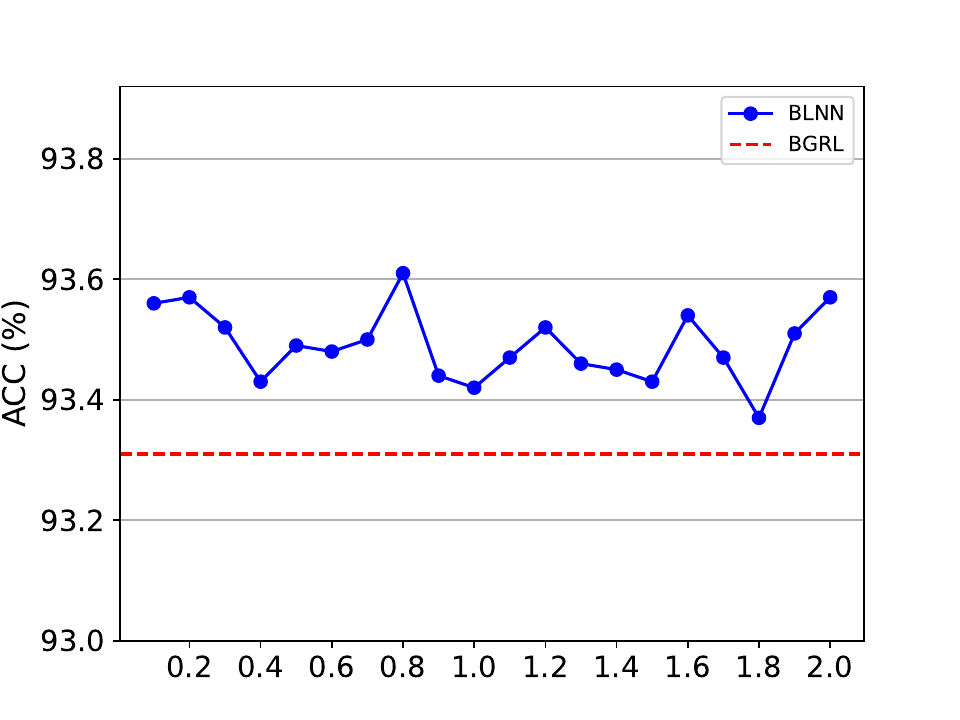}}
    \caption{Visualization of the impact of $\tau$ on node classification.}
    \label{Fig: Temperature}
\end{figure}

\noindent\textbf{Hyperparameter Analysis (RQ4).}\quad
We investigate the impact of the temperature $\tau$ in Eq. (\ref{Eq: Attention Weighting}) on node classification by varying $\tau$ from $0.1$ to $2.0$ in increments of $0.1$. Figure \ref{Fig: Temperature} presents the ACC scores on Photo, Computer and CS. It is observed that, our BLNN almost always achieves better performance than BGRL with respect to different $\tau$. In general, BLNN exhibits robustness to the temperature $\tau$. Analysis for BGRL-related hyperparameters can be found in the original BGRL paper \cite{BGRL}.

\noindent\textbf{Visualization and Compactness of Representations (RQ5).}\quad
To gain a more intuitive insight into node representations, we provide the t-SNE \cite{t-SNE} visualizations of the raw features and representations learned by BGRL and BLNN, along with intra-class compactness score on Computer. The intra-class compactness score is defined as the mean cosine similarity among all intra-class node pairs (the formula can be found in Appendix A.4). As shown in Figure \ref{Fig: t-SNE and Compactness}, the representations learned by BLNN exhibit higher intra-class compactness, thus underscoring the effectiveness of mining positive node-neighbor pairs.

\begin{figure}[h]
    \centering
    \subfigure[Raw$(0.4046)$]{\includegraphics[width=0.325\linewidth]{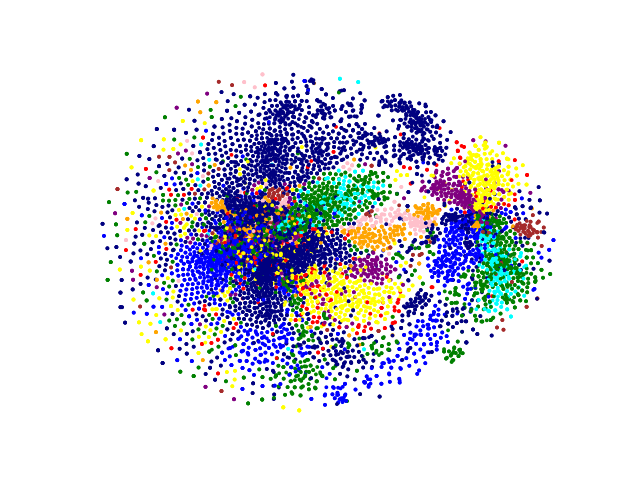}}  
    \subfigure[BGRL$(0.6733)$]{\includegraphics[width=0.325\linewidth]{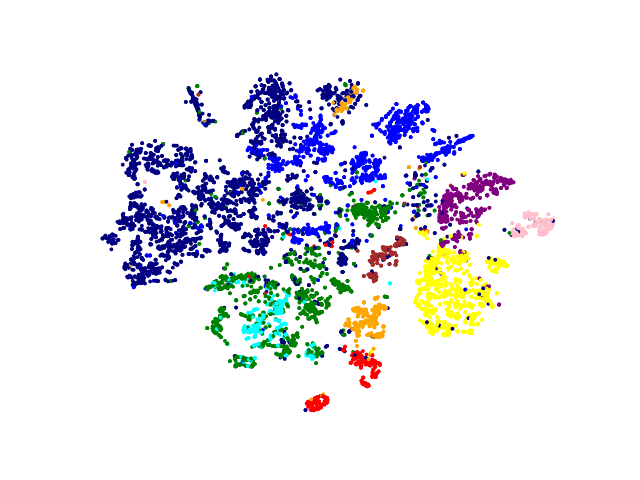}}
    \subfigure[BLNN$(0.7015)$]{\includegraphics[width=0.325\linewidth]{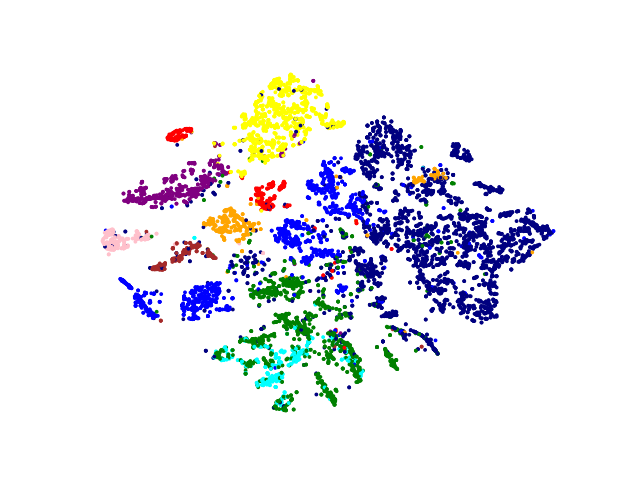}}
    \caption{t-SNE visualization and intra-class compactness of node representations on Computer. `$(*)$' indicates the mean intra-class pair-wise cosine similarity.
    }
    \label{Fig: t-SNE and Compactness}
\end{figure}

\section{Conclusion}
In this paper, we introduce Bootstrap Latents of Nodes and Neighbors (BLNN). Our proposal is motivated by the empirical observation that introducing ground-truth positive node-neighbor pairs can yield significant improvements for BGRL. We thus expand the positive pair set with node-neighbor pairs and propose a cross-attention module to weight the contribution of each neighbor to loss computations. This module prioritizes higher weights for ground-truth positive node-neighbor pairs compared to false positive node-neighbor pairs, thereby alleviating class collision resulting from the alignment of false node-neighbor pairs. Extensive experiments demonstrate that our BLNN effectively improves the intra-class compactness of learned representations, establishing its state-of-the-art performance in three downstream tasks across five benchmark datasets.

\section*{Acknowledgments}
This work is partially supported by the National Key Research and Development Program of China (2021YFB1715600), and the National Natural Science Foundation of China (62306137).

%
% ---- Bibliography ----
%
% BibTeX users should specify bibliography style 'splncs04'.
% References will then be sorted and formatted in the correct style.
%

\bibliographystyle{splncs04}
\bibliography{references}

\appendix
\section{Experiments}
\subsection{Datasets}
We evaluate our model on five representative datasets: WikiCS, Photo, Computer, CS and Physics. Their brief introductions are as follows:
\begin{itemize}
    \item \textbf{WikiCS} \cite{WikiCS} is a reference network constructed from Wikipedia. It comprises nodes corresponding to articles in the field of Computer Science, where edges are derived from hyperlinks. The dataset includes 10 distinct classes representing various branches within the field. The node features are computed as the average GloVe word embeddings of the respective articles.
    
    \item \textbf{Photo} and \textbf{Computer} \cite{Amazon} are networks constructed from Amazon's co-purchase relationships. Nodes represent goods, and edges indicate frequent co-purchases between goods. The node features are represented by bag-of-words encoding of product reviews, and class labels are assigned based on the respective product categories.

    \item \textbf{CS} and \textbf{Physics} \cite{Amazon} are co-authorship networks based on the Microsoft Academic Graph. Here, nodes are authors, that are connected by an edge if they co-authored a paper; node features represent paper keywords for each author’s papers, and class labels indicate most active fields of study for each author.
\end{itemize}
For all datasets, we use the processed version provided by PyTorch Geometric Library \cite{PyG}. All datasets are public available and do not have licenses.

\subsection{Baselines}
In this subsection, we give brief introductions of the baselines used in the paper which are not described in the main paper due to the space constraint.

\begin{itemize}
    \item \textbf{GCN} \cite{GCN} and \textbf{GAT} \cite{GAT} are two popular supervised graph neural networks that exploit structural information, raw node features, and node labels from the training set.

    \item \textbf{DGI} \cite{DGI} maximizes the mutual information between node representations and graph summary.

    \item \textbf{MVGRL} \cite{MVGRL} maximizes the mutual information between the cross-view representations of nodes and graphs using graph diffusion.

    \item \textbf{GRACE} \cite{GRACE} performs graph augmentation on the input graph and considers node-node level contrast on both inter-view and intra-view levels.

    \item \textbf{GCA} \cite{GCA} extends GRACE with adaptive augmentation that incorporates various priors for topological and semantic aspects of the graph.

    \item \textbf{AF-GCL} \cite{AF-GCL} is an augmentation-free graph contrastive learning method, wherein the self supervision signal is constructed based on the aggregated features.

    \item \textbf{COSTA} \cite{COSTA} proposes covariance-preserving feature augmentation to overcome the bias issue introduced by the topology graph augmentation in graph contrastive learning.

    \item \textbf{FastGCL} \cite{FastGCL} contrasts weighted-aggregated and non-aggregated neighborhood information, rather than disturbing the graph topology and node attributes, to achieve faster training and convergence speeds.

    \item \textbf{gCooL} \cite{gCooL} extends GRACE by jointly learning the community partition and node representations in an end-to-end fashion, thereby directly leveraging the inherent community structure within a graph.

    \item \textbf{ProGCL} \cite{ProGCL} extends GRACE by leveraging hard negtive samples via Expectation Maximization to fit the observed node-level similarity distribution. We adopt the ProGCL-weight version as no synthesis of new nodes is leveraged.

    \item \textbf{CGKS} \cite{CGKS} preserves diverse hierarchical information through graph coarsening and facilitates cross-scale information interactions among different coarse graphs.
 
    \item \textbf{CCA-SSG} \cite{CCASSG} leverages classical Canonical Correlation Analysis to formulate a feature-level objective which can discard augmentation-variant information and prevent dimensional collapse.

    \item \textbf{G-BT} \cite{G-BT} utilizes a cross-correlation-based loss function instead of negative samples, which enjoys fewer hyperparameters and significantly reduced computation time.

    \item \textbf{AFGRL} \cite{AFGRL} extends BGRL by creating an alternative graph view through the discovery of nodes sharing both local structural information and global semantics with the original graph.

    \item \textbf{GraphMAE} \cite{GraphMAE} is a masked graph auto-encoder that focuses on feature reconstruction with both a masking strategy and scaled cosine error.

    \item \textbf{BGRL} \cite{BGRL} adopts asymmetrical BYOL \cite{BYOL} structure to align node-itself pairs without relying on negative samples, thus avoiding a quadratic bottleneck and class collision.
\end{itemize}

\subsection{Graph Augmentation}
We employ two graph data augmentation strategies designed to enhance graph attributes and topology information, respectively. They are widely used in graph self-supervised learning \cite{GRACE,CCASSG,BGRL}.

\textbf{Feature Masking.}\quad
We randomly select a portion of the node features' dimensions and mask their elements with zeros. Formally, we first sample a random vector $\boldsymbol{\widetilde{m}} \in \{ 0, 1 \}^F$, where each dimension is drawn from a Bernoulli distribution with probability $1 - p_m$, i.e., $\widetilde{m}_i \sim \mathcal{B}(1 - p_m), \forall i$. Then, the masked node features $\widetilde {\boldsymbol{X}}$ are computed by $\parallel_{i=1}^N \boldsymbol{x}_i \odot \boldsymbol{\widetilde{m}}$, where $\odot$ denotes the Hadamard product and $\parallel$ represents the stack operation (i.e., concatenating a sequence of vectors along a new dimension).

\textbf{Edge Dropping.}\quad
In addition to feature masking, we stochastically drop a certain fraction of edges from the original graph. Formally, since we only remove existing edges, we first sample a random masking matrix $\boldsymbol{\widetilde{M}} \in \{ 0, 1 \}^{N \times N}$, with entries drawn from a Bernoulli distribution $\boldsymbol{\widetilde{M}}_{i,j} \sim \mathcal{B}(1 - p_d)$ if $\boldsymbol{A}_{i,j} = 1$ for the original graph, and $\boldsymbol{\widetilde{M}}_{i,j} = 0$ otherwise. Here, $p_d$ represents the probability of each edge being dropped. The corrupted adjacency matrix can then be computed as $\boldsymbol{\widetilde{A}} = \boldsymbol{A} \odot \boldsymbol{\widetilde{M}}$.

We jointly utilize these two methods to generate graph views. And the hyperparameter settings for graph augmentations are the same as those in BGRL \cite{BGRL}.

\subsection{Formulas of Metrics}
We denote the ground-truth class labels as$\boldsymbol{Y}=[y_i]_{i=1}^{n}$ and the labels predicted by a classifier or clustering model as $\boldsymbol{\hat{Y}}=[\hat{y}_i]_{i=1}^{n}$.

\textbf{Accuracy} is determined as the proportion of correct predictions:
\begin{equation}
    \operatorname{ACC} = \frac{1}{n} \sum_{i=1}^n \mathbb{I}(y_i=\hat{y}_i),
\end{equation}
where $\mathbb{I}$ denotes the indicator function.

\textbf{Normalized Mutual Information} (NMI) measures the mutual information between the true class labels and the cluster assignments, normalized by the entropy of the class labels and the entropy of the cluster assignments. It is defined as:
\begin{equation}
    \operatorname{NMI} = \frac{2 I(\boldsymbol{Y};\boldsymbol{\hat{Y}})}{H(\boldsymbol{Y}) + H(\boldsymbol{\hat{Y}})},
\end{equation}
where $I(\cdot)$ is the mutual information, and $H(\cdot)$ is the entropy.

\textbf{Homogeneity} measures the degree to which each cluster contains only members of a single class:
\begin{equation}
    \operatorname{Homo.} = 1 - \frac{H(\boldsymbol{Y}|\boldsymbol{\hat{\boldsymbol{Y}}})}{H(\boldsymbol{Y})}.
\end{equation}

\textbf{S@}$\boldsymbol{k}$ denotes the percentage of the top $k$ neighbors that belong to
the same class. It is defined as:
\begin{equation}
    \operatorname{S@}k = \frac{1}{nk} \sum_{i=1}^n \sum_{j \in \mathcal{N}_k(i)} \mathbb{I}(y_i = y_j),
\end{equation}
where $\mathcal{N}_k(i)$ denotes the $k$ nearest neighbor set of $i$.

\textbf{Intra-class Compactness} of node representations is defined as the mean cosine similarity among all intra-class node pairs:
\begin{equation}
    \mathcal{C} = \frac{1}{K} \sum_{l=1}^{K} \frac{1}{|\boldsymbol{Y}=l|} \sum_{y_i=y_j=l}^{i \not= j} \operatorname{cos}(\boldsymbol{h}_i, \boldsymbol{h}_j),
\end{equation}
where $K$ is the number of unique classes, $|\boldsymbol{Y}=l|$ is the number of nodes belonging to class $l$, and $\operatorname{cos}(\boldsymbol{h}_i, \boldsymbol{h}_j)$ is the cosine similarity between node representations $\boldsymbol{h}_i, \boldsymbol{h}_j$.

\subsection{Implementation Details}
Since our BLNN is derived from BGRL, we implement BLNN based on the official code\footnote{\url{https://github.com/nerdslab/bgrl}} of BGRL. To ensure a fair comparison, all BGRL-related hyperparameters are the same as those specified in the original BGRL paper. Specially, we use the AdamW optimizer \cite{AdamW} with weight decay set to $10^{-5}$, and all models initialized using Glorot initialization \cite{Glorot}. The encoders $f_\theta, f_\phi$ are implemented as GCN \cite{GCN} and the predictor $p_\theta$ used to predict the embedding of nodes across views is fixed to be a Multilayer Perceptron (MLP) with a single hidden layer. The decay rate $t$ controlling the rate of updates of the target parameters $\phi$ is initialized to $0.99$ and gradually increased to $1.0$ over the course of training following a cosine schedule. We perform a grid-search on the introduced temperature hyperparameter $\tau$. Other model architecture and training details can be found in the original BGRL paper \cite{BGRL}. All experiments are conducted on a 32GB V100 GPU. Our implementation code is available at \url{https://github.com/Cloudy1225/BLNN}.

\begin{table}[h]
\caption{Comparison with HomoGCL on node classification. The BGRL* and HomoGCL results are taken from the original HomoGCL paper, with the BGRL* results reproduced by HomoGCL's authors.}
    \label{Tab: Comparison with HomoGCL}
    \centering
    \begin{tabular}{lcccc}
        \toprule
        ~             & BGRL* & HomoGCL & BLNN & BGRL    \\
        \midrule
        Photo         & 92.80 & 93.53  & 93.54 & 90.17    \\
        Computer      & 88.23 & 90.01  & 91.02 & 90.34    \\
        \bottomrule
    \end{tabular}
\end{table}

\subsection{Comparison with HomoGCL}
We observed that a peer study \cite{HomoGCL}, called HomoGCL, shares certain similarities with our method. HomoGCL leverages homophily by estimating the probability of neighbor nodes being positive via Gaussian Mixture Model. It then softly aligns the representations of node-neighbor pairs and directly aligns the cluster assignment vectors of node-neighbor pairs. We provide node classification results in Table \ref{Tab: Comparison with HomoGCL}. The BGRL* and HomoGCL results are taken from the original HomoGCL paper, with the BGRL* results reproduced by HomoGCL's authors. We can find that our BLNN exhibits nearly identical performance to HomoGCL on Photo and demonstrates a substantial improvement on Computer. Additionally, HomoGCL requires performing time-consuming K-means clustering on the entire set of node representations to estimate cluster assignments. Finally, we express our gratitude to the authors of HomoGCL for their outstanding contributions to the graph self-supervised learning community.

\end{document}